\documentclass[pmlr]{jmlr-mod}

\RequirePackage{graphicx}
\usepackage{booktabs}
\usepackage{longtable}

\providecommand{\abs}[1]{\ensuremath{\lvert#1\rvert}}

\usepackage{subcaption}
\usepackage{multirow}

\usepackage[load-configurations=version-1]{siunitx} 
\jmlryear{2025}
\jmlrworkshop{}

\title[FMs for EHRs: representation dynamics and transferability]{Foundation models for electronic health records: representation dynamics and transferability}

\author{\Name{Michael C. Burkhart}
       \Email{burkh4rt@uchicago.edu}\\ 
       \addr Center for Computational Medicine \& Clinical AI, Department of Medicine \\
       University of Chicago\\
       Chicago, IL, USA
       \AND
       \Name{Bashar Ramadan}
       \Email{basharramadan@uchicago.edu}\\ 
       \addr Section of Hospital Medicine, Department of Medicine \\
       University of Chicago\\
       Chicago, IL, USA
       \AND
       \Name{Zewei ``Whiskey'' Liao}
       \Email{wliao0504@uchicago.edu}\\ 
       \Name{Kaveri Chhikara}
       \Email{kaveri@uchicago.edu}\\ 
       \addr Center for Computational Medicine \& Clinical AI, Department of Medicine \\
       University of Chicago\\
       Chicago, IL, USA
       \AND
       \Name{Juan C. Rojas}
       \Email{juan\_rojas@rush.edu}\\ 
       \addr Department of Internal Medicine \\
       Rush University\\
       Chicago, IL, USA
       \AND
       \Name{William F. Parker}
       \Email{wparker@uchicago.edu}\\ 
       \Name{Brett K. Beaulieu-Jones}
       \Email{beaulieujones@uchicago.edu}\\ 
       \addr Center for Computational Medicine \& Clinical AI, Department of Medicine \\
       University of Chicago\\
       Chicago, IL, USA
       }

\begin{document}

\maketitle

\begin{abstract}
	Foundation models (FMs) trained on electronic health records (EHRs) have shown strong performance on a range of clinical prediction tasks. However, adapting these models to local health systems remains challenging due to limited data availability and resource constraints. In this study, we investigated what these models learn and evaluated the transferability of an FM trained on MIMIC-IV to an institutional EHR dataset at the University of Chicago Medical Center. We assessed their ability to identify outlier patients and examined representation-space patient trajectories in relation to future clinical outcomes. We also evaluated the performance of supervised fine-tuned classifiers on both source and target datasets. Our findings offer insights into the adaptability of FMs across different healthcare systems, highlight considerations for their effective implementation, and provide an empirical analysis of the underlying factors that contribute to their predictive performance.
\end{abstract}

\section{Introduction}

Large language model (LLM) architectures trained on sequences of tokenized electronic healthcare records (EHRs) have proven to be excellent foundation models (FMs) for an array of prognostic tasks. In a standard workflow, tables of clinical data are converted into patient- or visit-level sequences of tokens~\citep{Ste21,McD23}. Foundation models are then trained to generate completions for these sequences using a self-supervised objective. These models can then directly predict outcomes of interest for a novel timeline (zero-shot learning), provide representations for each sequence that can be used with off-the-shelf classification or regression models, or be further fine-tuned in a supervised fashion to predict specific outcomes at a lower computational cost than the initial training. Outcomes commonly considered include inpatient mortality or mortality within a certain time horizon, ICU transfer or readmission, long length of stay, the results of lab tests, the assignment of diagnoses, and imaging findings~\citep{Wor23}. In the near future, these models could improve patient risk stratification, suggest targeted interventions, and allow hospitals to better predict resource requirements \citep{rajpurkar2022ai, beam2016translating, yu2019framing}. In the longer term, these models could help us to better understand the etiology and progression of disease \citep{ching2018opportunities, singhal2023opportunities}.

While established common data models (CDMs) like the Observational Medical Outcomes Partnership (OMOP)~\citep{stang2010advancing}, i2b2 \citep{murphy2010serving}, the FDA's Sentinel Initiative CDM \citep{racoosin2012fda}, and PCORnet's CDM \citep{fleurence2014launching} aim to standardize EHR data for broad research use~\citep{garza2016evaluating}, they often face challenges in representing the granular, high-frequency data characteristic of critical care settings. For instance, detailed ventilator settings, continuous infusion rates, and specific ICU interventions may be captured inconsistently or lack the necessary detail in these general CDMs, hindering multi-center critical care research. The Common Longitudinal ICU data Format \citep[CLIF:][]{Roj25} was developed to address this gap, offering a standardized structure specifically designed for the complexities of ICU data. By focusing on a minimum set of essential Common ICU Data Elements (mCIDE) within a clinically intuitive, longitudinal format, CLIF complements broader CDMs and provides a more suitable foundation for research requiring detailed critical care information, facilitating reproducible studies across diverse healthcare systems.

Previous research suggests that representations learned from FMs provide improved robustness to distribution shift for a range of classification tasks~\citep{Guo23} and can be successfully adapted to new sites with additional training~\citep{Guo24}. In this paper, we focus on understanding the dynamics of FM-learned representations and assessing the transferability of these representations.

\subsection{Generalizable Insights}
FMs on EHRs have achieved impressive results on a variety of predictive tasks. Many of the successful models have extracted FM-derived representations of patient timelines and used these as features to train outcome-specific classifiers. In this paper, we show that out-of-the-box performance of classifiers built in this way can seriously degrade when transferred to a new dataset, and explore some ways to mitigate this performance decline. As models trained on MIMIC are transferred to new hospital systems with different clinical practices and more diverse populations, we would expect practitioners to face similar problems with model transfer. Our results suggest that fine-tuning models can help mitigate performance degradation during model transfer, even when a substantial fraction of the new dataset is viewed as anomalous in reference to the original data.

\section{Related Work}
Early applications of deep learning for predicting clinical outcomes from sequential EHR data leveraged recurrent neural networks (RNNs) to model patient trajectories. \citet{lipton2017learningdiagnoselstmrecurrent} investigated the use of Long Short-Term Memory (LSTM) networks for multilabel classification of diagnoses from multivariate clinical time series, highlighting the effectiveness of RNNs in handling irregularly sampled and missing data in EHRs. \citet{choi2016doctoraipredictingclinical} employed RNNs to forecast future diagnoses and medication prescriptions based on patients' historical records, demonstrating the potential of deep learning in capturing temporal patterns within EHR data. Similarly, \citet{beaulieu2018mapping} used RNNs to predict critical care outcomes and demonstrated value in clinically-appropriate groupings and representations of critical care data for deep learning applications.

\citet{rajkomar2018scalable} applied deep learning approaches to clinical data represented in an interoperable format (FHIR) and scaled up the approaches to 216,221 patients using over 200,000 GPU hours. This was followed by work investigating whether these models were learning from the patient state or from clinical behavior in order to assess these types of models in terms of their ability to be used for individualized clinical decision making \citep{beaulieu2021machine}.

Approaches shifted from recurrent neural networks to transformers on EHR representations beginning with variations on BERT~\citep{Dev19}, including BEHRT~\citep{Li20} and Med-BERT~\citep{Ras21}. Foresight~\citep{Kra24} and ETHOS~\citep{Ren24,Ren25} both use generative pretrained transformer~\citep[GPT:][]{Rad18} architectures. \citet{Wor23b} provide a more detailed review of FMs for EHRs. More recently, Mamba~\citep{Gu24}, a selective state-space model, has found applications in ClinicalMamba~\citep{Yan24} and EHRMamba~\citep{Fal24}. Comparisons of these architectures indicate similar performance for short contexts and an advantage to Mamba for longer contexts~\citep{Wor25}. Applications include anomaly detection~\citep{Niu24}, disease-specific adaptations that model a prodromal time window~\citep[TRADE:][]{Zhu24}, and time-to-event models for improved predictive precision~\citep[MOTOR:][]{Ste24}.

\section{Methods}

\subsection{Training}
\label{ss:methods:original_training}
\begin{figure}[ht]
	\centering
	\subfigure[Self-supervised training with packing]{
		\centering
		\includegraphics[width=0.43\textwidth]{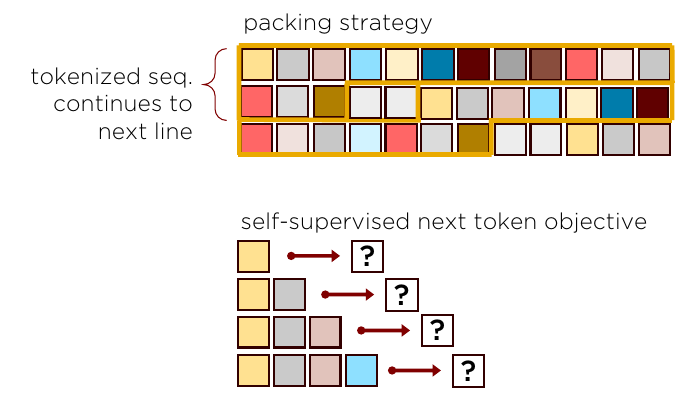}
	}
	\hspace{10pt}
	\subfigure[Fine-tuning for a specific outcome]{
		\centering
		\includegraphics[width=0.43\textwidth]{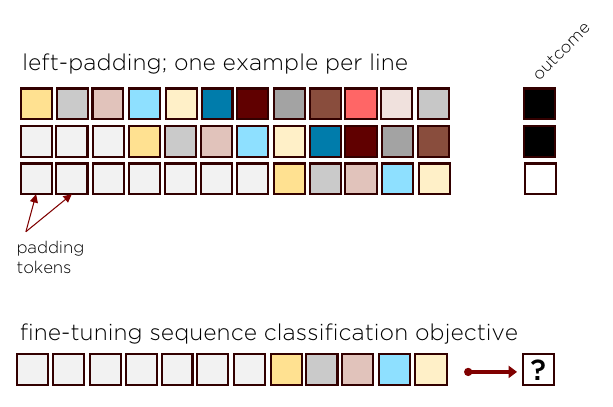}
	}
	\caption{In (a), our initial training process packed sequences together, allowing one sequence to bleed into the next example within a batch. The dark goldenrod boundary outlines tokens corresponding to two individual hospitalization events. We insert a variable number of padding tokens between sequences to expose the model to padding. For the initial training, the model attempted to predict the next token in a sequence given the previous tokens (`context'). In (b), we performed supervised fine-tuning with left-padded sequences. Each hospitalization event (truncated at 24 hours) occupies a single training instance and is paired with its associated subsequent outcome. In this way, fine-tuning is outcome-specific.}
	\label{fig:training_sft}
\end{figure}

We used the Llama-3.2 1B-parameter architecture~\citep{Gra24} and trained with the standard self-supervised objective of next token prediction. Patient sequences (examples) were fed into the model using a packing strategy, with a random number of padding tokens inserted between each timeline to expose the model to padding tokens during training. We used tree-structured Parzen estimators to tune the learning rate and effective batch size~\citep{Aki19}. See Figure~\ref{fig:training_sft}(a) for a visualization.

\subsection{Representation-based classifiers}
\label{ss:rep_clsfrs}
After training completed, we selected the sub-sequences of timelines corresponding to events within the first 24 hours of admission (truncating at 1024 tokens when necessary). We then used the trained models to extract a latent representation for these sequences. For the MIMIC training and validation sets, we fit logistic regression models for our outcomes of interest and then performed inference using these models on the MIMIC and University of Chicago Medical Center (UCMC) sets.

\subsection{Outlier detection}
\label{ss:out}
We trained an Isolation Forest~\citep{Liu08} on the 24-hour representations to detect outliers on the MIMIC training set and applied it to the other datasets. The forest is an ensemble of $n=100$ trees, each of which iteratively partitions a subset $D_i$ of the data by randomly selecting a dimension and then sampling a uniform split along that dimension. For a data point $x \in D_i$, the number of splits required to isolate that point from the remainder of $D_i$ in the $i$th tree is denoted $h_i(x)$. The underlying idea is that, on average, $h_i(x)$ is inversely related to the anomalousness of $x$. Explicitly, the anomaly score $s(x)$ associated to $x$ is given by $s(x)= 2^{-\tfrac{1}{\abs{I(x)}} \sum_{i \in I(x)} h_i(x)/c}$, where $I(x)= \{i: x \in D_i\}$ and $c>0$ depends on the size of the dataset, with higher values corresponding to likely outliers.

\subsection{Fine-tuned classifiers}
\label{ss:ft}

\subsubsection{On complete sequences}
\label{ss:sft}
We also took models trained as described in~\S\ref{ss:methods:original_training}, added a classification head, and finetuned them to predict our four outcomes of interest: in-hospital mortality, long length of stay ($\ge$7 days), ICU admission after 24 hours, and IMV event after 24 hours. Finetuning was completed at a much reduced learning rate and used a left-padding strategy for sequence preparation. For fine-tuning on MIMIC, the learning rate was fixed at $2\cdot 10^{-5}$. See Figure~\ref{fig:training_sft}(b) for a visualization.

\subsubsection{For partial sequence prediction}
Partial sequence prediction allows for the analysis of prediction changes over the time that a patient is in the hospital. Preliminary work indicated that, as some truncations are impossible according to our grammar (for example, a decile token should always immediately follow a lab token in any valid sequence), models trained on full sequences displayed high variability on partial sequence prediction tasks. To mitigate these issues, we fine-tuned models for partial sequence prediction, by forming sequences truncated uniformly at random and supplying them to the classifier with their corresponding labels.

\subsubsection{Additional Local fine-tuning}
\label{ss:sft_local}
Up to this point, all models were trained strictly using the MIMIC training and validation sets, and not adapted to the UCMC dataset in any way. To determine if performance could be further improved by continuing to train on UCMC data, we took models fine-tuned on MIMIC and performed the same fine-tuning process on UCMC training data, with performance monitoring on the UCMC validation set. We selected hyperparameters for local fine-tuning using tree-structured Parzen estimators.

\subsection{Models for partial sequence classification}
\label{ss:partial_seq_clsf}
To better understand how model predictions change as tokens are added, we compared multiple models for partial sequence prediction. In this task, we predict in-hospital mortality for a patient given the first $i$ tokens in their timeline up to the 24-hour cutoff point, for each $1\leq i \leq n$, where $n$ denotes the full length of their sequence. We compared the two models trained as described in~\S\ref{ss:ft} with an adapted logistic regression-based approach trained as follows. For each patient in the MIMIC training set, we extracted a state representation from the first $i$ steps, where $i$ was chosen uniformly at random from $\{1,\dotsc,n\}$, and paired it with a flag corresponding to the patient's mortality outcome.

\section{Cohort}

\subsection{Cohort Selection}
The study population was adults (age 18 or older) hospitalized at the Beth Israel Deaconess Medical Center between 2008–2019~\citep[MIMIC-IV-3.1:][]{Joh23} and the University of Chicago Medical Center (UCMC) from March 2020 to March 2022. We formatted EHR data from each health system into the CLIF standard~\citep{Roj25}. The MIMIC patients were partitioned intro training, validation, and test sets at a 70\%-10\%-20\% rate, according to the randomized time of their first hospitalization event, with training patients coming first, followed by validation and then test. We then collected each hospitalization event for patients in a given set. In this way, hospitalization records in the test set corresponded to patients with no hospitalization events in the training or validation sets. The UCMC data was primarily used as a held-out test set; with the finetuning in \S\ref{ss:sft_local} being the only exception. For this reason, we partitioned the UCMC patients intro training, validation, and test sets at a 5\%-5\%-90\% rate in the same manner as used for MIMIC, and formed hospitalization records in an analogous manner. For demographic details on the MIMIC and UCMC datasets, please consult tables~\ref{tbl:summary_mimic} and~\ref{tbl:summary_internal}, respectively.

\begin{table}[ht]
	\centering
	\scalebox{0.85}{
		\begin{tabular}{lrr>{\bfseries}rrr>{\bfseries}rrr>{\bfseries}r}
			\toprule
			                     & \multicolumn{3}{c}{\textbf{train}} & \multicolumn{3}{c}{\textbf{val}} & \multicolumn{3}{c}{\textbf{test}}                                                           \\
			\cmidrule(lr){2-4} \cmidrule(lr){5-7} \cmidrule(lr){8-10}
			                     & inliers                            & outliers                         & all                               & inliers & outliers & all   & inliers & outliers & all   \\
			\midrule
			count                & 261736                             & 27941                            & 289677                            & 38799   & 4058     & 42857 & 82429   & 7811     & 90240 \\
			timeline len. (@24h) & 62.9                               & 465.2                            & 101.7                             & 62.0    & 459.7    & 99.7  & 61.8    & 467.1    & 96.9  \\ \midrule
			age (avg.)           & 60.0                               & 63.6                             & 60.4                              & 60.6    & 63.9     & 60.9  & 61.0    & 63.7     & 61.3  \\
			fraction female      & 0.536                              & 0.429                            & 0.526                             & 0.537   & 0.431    & 0.527 & 0.543   & 0.435    & 0.534 \\
			-- African American  & 0.157                              & 0.099                            & 0.151                             & 0.158   & 0.107    & 0.153 & 0.172   & 0.113    & 0.167 \\
			-- Asian             & 0.037                              & 0.030                            & 0.036                             & 0.038   & 0.028    & 0.037 & 0.034   & 0.034    & 0.034 \\
			-- Caucasian         & 0.686                              & 0.672                            & 0.684                             & 0.683   & 0.674    & 0.682 & 0.682   & 0.674    & 0.682 \\
			-- Native American   & 0.003                              & 0.002                            & 0.003                             & 0.002   & 0.002    & 0.002 & 0.002   & 0.001    & 0.002 \\
			-- Pacific Islander  & 0.001                              & 0.001                            & 0.001                             & 0.000   & 0.001    & 0.000 & 0.001   & 0.001    & 0.001 \\
			-- Unknown/Other     & 0.117                              & 0.196                            & 0.125                             & 0.119   & 0.188    & 0.125 & 0.108   & 0.177    & 0.114 \\
			-- Hispanic          & 0.057                              & 0.037                            & 0.055                             & 0.063   & 0.041    & 0.061 & 0.056   & 0.040    & 0.055 \\ \midrule
			inhospital mortality & 0.016                              & 0.105                            & 0.024                             & 0.016   & 0.098    & 0.023 & 0.014   & 0.107    & 0.022 \\
			long length of stay  & 0.208                              & 0.434                            & 0.230                             & 0.198   & 0.425    & 0.220 & 0.198   & 0.449    & 0.220 \\
			ICU (w/in 24h)       & 0.080                              & 0.869                            & 0.156                             & 0.076   & 0.866    & 0.151 & 0.076   & 0.871    & 0.145 \\
			ICU (any)            & 0.124                              & 0.886                            & 0.198                             & 0.119   & 0.881    & 0.191 & 0.117   & 0.889    & 0.184 \\
			IMV (w/in 24h)       & 0.009                              & 0.406                            & 0.048                             & 0.009   & 0.387    & 0.045 & 0.009   & 0.396    & 0.042 \\
			IMV (any)            & 0.033                              & 0.474                            & 0.076                             & 0.031   & 0.447    & 0.071 & 0.031   & 0.466    & 0.068 \\
			\bottomrule
		\end{tabular}
	}
	\caption{Summary of MIMIC data splits and outlier status by demographics and outcomes.}
	\label{tbl:summary_mimic}
\end{table}

\begin{table}[ht]
	\centering
	\scalebox{0.85}{
		\begin{tabular}{lrr>{\bfseries}rrr>{\bfseries}rrr>{\bfseries}r}
			\toprule
			                     & \multicolumn{3}{c}{\textbf{train}} & \multicolumn{3}{c}{\textbf{val}} & \multicolumn{3}{c}{\textbf{test}}                                                           \\
			\cmidrule(lr){2-4} \cmidrule(lr){5-7} \cmidrule(lr){8-10}
			                     & inliers                            & outliers                         & all                               & inliers & outliers & all   & inliers & outliers & all   \\
			\midrule
			count                & 3950                               & 3047                             & 6997                              & 2425    & 2145     & 4570  & 27917   & 26074    & 53991 \\
			timeline len. (@24h) & 342.2                              & 430.9                            & 380.8                             & 372.7   & 470.4    & 418.6 & 352.0   & 462.7    & 405.4 \\ \midrule
			age (avg.)           & 52.3                               & 54.4                             & 53.2                              & 54.9    & 56.2     & 55.5  & 54.4    & 55.4     & 54.9  \\
			fraction female      & 0.574                              & 0.569                            & 0.572                             & 0.515   & 0.533    & 0.523 & 0.570   & 0.552    & 0.561 \\
			-- African American  & 0.808                              & 0.786                            & 0.798                             & 0.776   & 0.767    & 0.772 & 0.662   & 0.643    & 0.653 \\
			-- Asian             & 0.013                              & 0.019                            & 0.016                             & 0.011   & 0.013    & 0.012 & 0.019   & 0.020    & 0.019 \\
			-- Caucasian         & 0.148                              & 0.160                            & 0.153                             & 0.167   & 0.170    & 0.169 & 0.261   & 0.271    & 0.266 \\
			-- Native American   & 0.000                              & 0.000                            & 0.000                             & 0.001   & 0.003    & 0.002 & 0.002   & 0.002    & 0.002 \\
			-- Pacific Islander  & 0.000                              & 0.000                            & 0.000                             & 0.000   & 0.001    & 0.001 & 0.001   & 0.001    & 0.001 \\
			-- Unknown/Other     & 0.031                              & 0.035                            & 0.033                             & 0.045   & 0.046    & 0.045 & 0.055   & 0.063    & 0.059 \\
			-- Hispanic          & 0.038                              & 0.044                            & 0.041                             & 0.035   & 0.041    & 0.038 & 0.056   & 0.061    & 0.059 \\ \midrule
			inhospital mortality & 0.013                              & 0.034                            & 0.022                             & 0.021   & 0.046    & 0.033 & 0.015   & 0.040    & 0.027 \\
			long length of stay  & 0.238                              & 0.292                            & 0.262                             & 0.259   & 0.315    & 0.285 & 0.226   & 0.283    & 0.254 \\
			ICU (w/in 24h)       & 0.064                              & 0.160                            & 0.105                             & 0.095   & 0.193    & 0.141 & 0.090   & 0.202    & 0.144 \\
			ICU (any)            & 0.102                              & 0.211                            & 0.149                             & 0.136   & 0.245    & 0.187 & 0.127   & 0.247    & 0.185 \\
			IMV (w/in 24h)       & 0.009                              & 0.067                            & 0.034                             & 0.016   & 0.074    & 0.043 & 0.014   & 0.086    & 0.049 \\
			IMV (any)            & 0.025                              & 0.092                            & 0.054                             & 0.034   & 0.111    & 0.070 & 0.033   & 0.117    & 0.074 \\
			\bottomrule
		\end{tabular}
	}
	\caption{Summary of UCMC data splits and outlier status by demographics and outcomes.}
	\label{tbl:summary_internal}
\end{table}

Our tokenization process operated on all nine of the tables available at the time of publication: \texttt{patient}, containing patient-level demographic data; \texttt{hospitalization}, with admission- and discharge-related data; \texttt{adt}, containing intra-hospital transfer data; \texttt{vitals}, in a standardized format; \texttt{patient\_assessments}; \texttt{respiratory support}, including records of invasive mechanical ventilation; \texttt{labs}; and \texttt{medication\_admin\_continuous}, containing data on continuously-administered medication.

\subsection{Data Extraction}

\begin{figure}[htb]
	\centering
	\subfigure[Category-value tokenization]{
		\centering
		\includegraphics[width=0.8\textwidth]{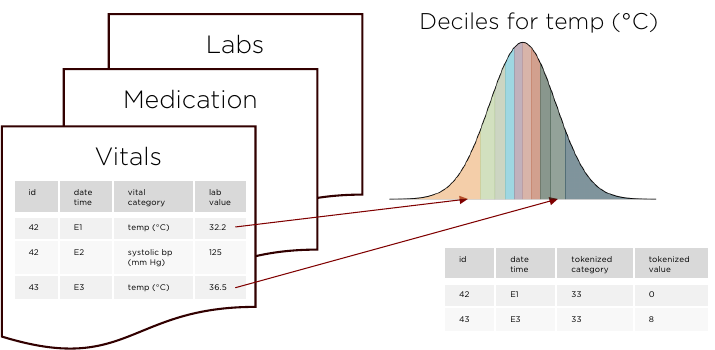}
		\label{fig:catval}
	}
	\hspace{10pt}
	\subfigure[Example tokenized timeline]{
		\centering
		\includegraphics[width=0.8\textwidth]{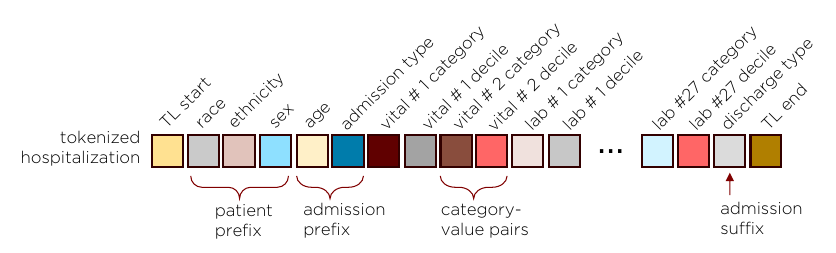}
	}
	\label{fig:timeline}
	\caption{The tokenization process converts information and events associated to a hosptialization into a sequence of integers. (a) Category-value tokenization iterates over all categories present in a table and learns deciles for the values within each category. In this example, we see how the vital corresponding to temperature in Celsius is assigned the label `33.' All measurements of temperature in the MIMIC training set are used to determine deciles for measurements within this category. For hospitalization 42, the tokens `33'  for this category and then `0' for the corresponding deciled measurement would be inserted into the timeline at `E1'. In (b), we see the anatomy of a basic timeline, starting with a start token, including some information about the patient, the admission, and then a series of measurements inserted in chronological order describing their visit, followed by a discharge token, and a token for timeline end.}
	\label{fig:tokenization}
\end{figure}

A hospitalization event was tokenized, or converted into sequences of integers, as follows. The first token always corresponds to timeline start token. The next three tokens contain patient-level demographic information on race, ethnicity, and sex. The following two tokens correspond to admission-specific information, namely patient age converted to a decile and admission type. Tokens corresponding to a variety of events for a hospitalization are then inserted in the same order in which these events occurred. Transfers are encoded with their CLIF location category. Labs are encoded with two tokens and inserted at the time results become available: one for the lab category, and a second corresponding to the deciled lab value in the training data within that category. We call this strategy, of tokenizing categories and binning their corresponding values according to the training value of the deciles, category-value tokenization.  A handful of other tables receive this type of tokenization: vitals and results according to vital category, medication and dosage by medication category, assessment and results by assessment category. Respiratory information is recorded at the beginning of respiratory support; the encoded information is mode category and device category. We include a token indicating if a patient is placed in a prone position. All hospitalization-related data is encoded this way and inserted in chronological order. Timelines then end with a token for discharge category and a dedicated timeline end token. We restricted to patients with stays of at least 24 hours. See Figure~\ref{fig:tokenization} for a visualization.

In order to evaluate the prognostic capability of our models, we also form a version of our timelines using only information available within 24 hours of admission. Many of the original papers on FMs for EHRs focused on long-term outcomes and considered timelines that spanned multiple years of patient data. There's been an increasing focus on predicting same-admission outcomes, like the outcomes we consider here. Predicting same-admission outcomes generally requires more care be taken to avoid information leakage. For example, we do not use ICD-10 diagnostic codes in our model, as these are finalized only after a patient is discharged. This is described in the MIMIC-IV publication \citep{Joh23} which states: ``the diagnoses\_icd table contains coded diagnoses representing the hospitalization as determined by trained professionals after reviewing signed patient notes.'' Our labs are inserted into timelines at the point of availability, and not when the sample was collected.
\subsection{Outcomes}
We considered four outcomes of clinical interest: (1) inpatient mortality, defined as patient death prior to discharge, (2) long length of stay, defined as more than 7 days passing between admission and discharge, (3) ICU admission, defined as transfer to an ICU ward as categorized in CLIF \citep{Roj25}, and (4) an Invasive Mechanical Ventilation (IMV) event, also as categorized in CLIF. For evaluating outcomes (3) \& (4), we restricted to patients who did not have that outcome within the first 24 hours of their stay.

\section{Results on Real Data}

\begin{figure}[ht]
	\centering
	\subfigure[all tokens]{
		\centering
		\includegraphics[width=0.43\textwidth]{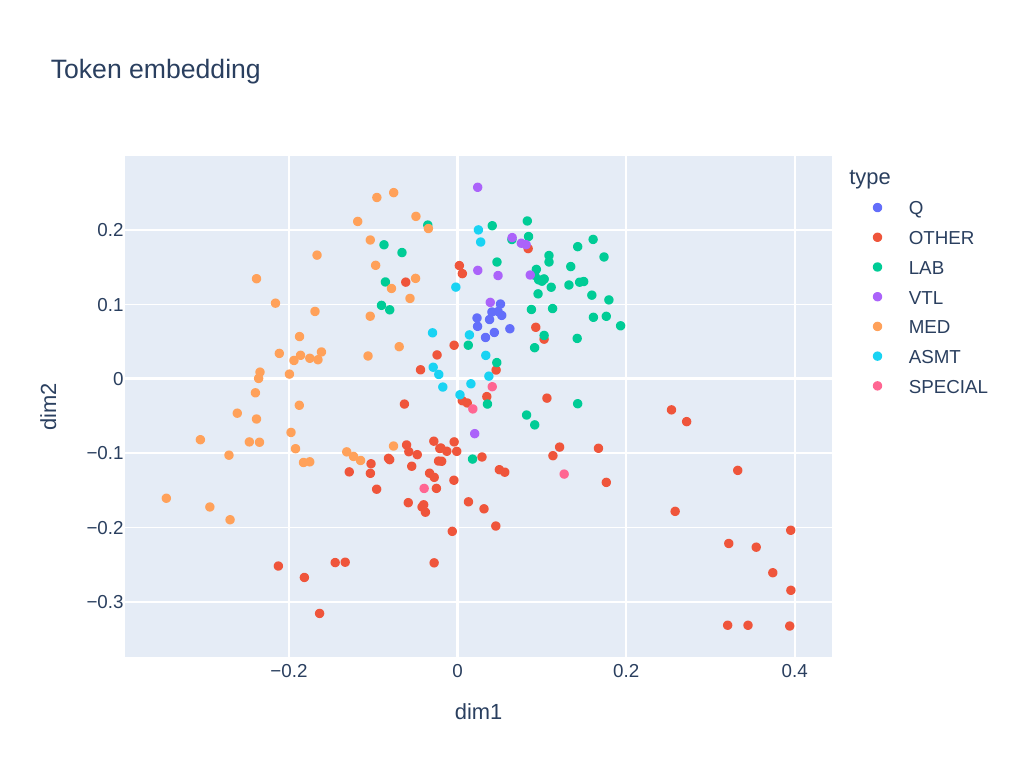}
		\label{fig:pca_all}
	}
	\hspace{10pt}
	\subfigure[decile tokens]{
		\centering
		\includegraphics[width=0.43\textwidth]{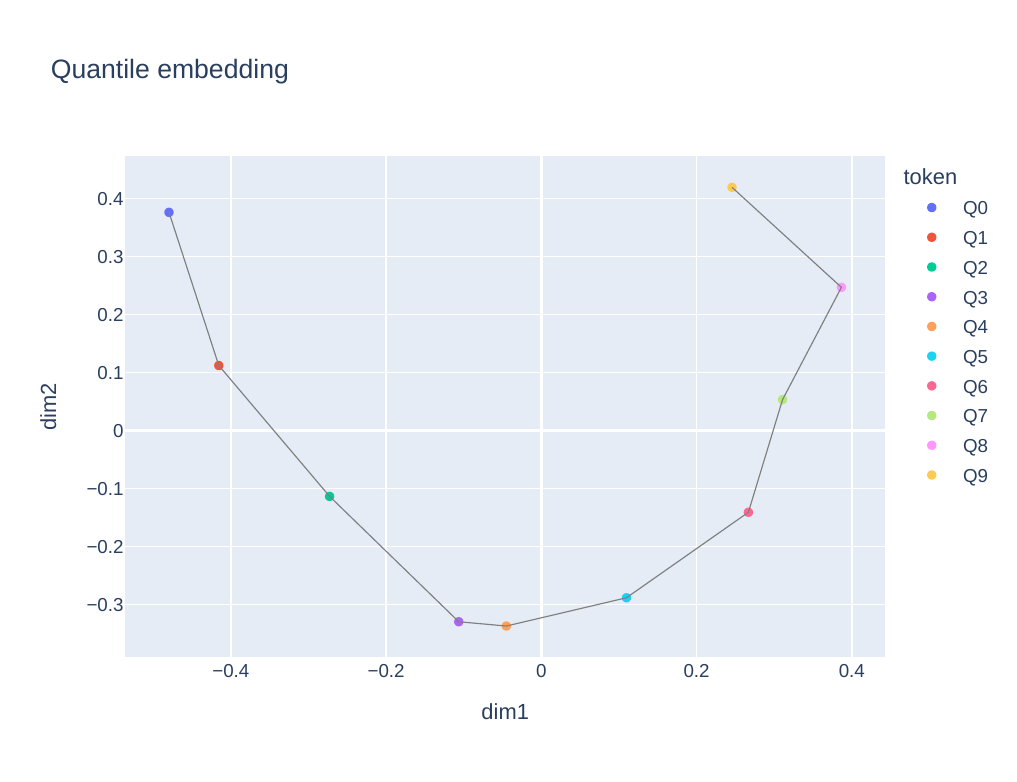}
		\label{fig:pca_10q}
	}
	\caption{We plot the first two PCA components for the embeddings corresponding to ({\it a}) all tokens, colored by token type and ({\it b}) the ten quantile tokens. In ({\it a}), we note that tokens corresponding to similar categories tend to be grouped within the embedding. For ({\it b}), we note that the model successfully learned the relative ordering of the deciles.}
	\label{fig:embeddings}
\end{figure}

Upon completion of training the baseline model, we plotted the first 2 PCA components of all token embeddings and decile embeddings (Figure~\ref{fig:embeddings}). We then extracted features for all data subsets and trained logistic regression models as described in~\S\ref{ss:rep_clsfrs}.

Performance, as given in Table~\ref{tbl:rep-based-rocs}, was acceptable in MIMIC, but lost a great deal of predictive power when transferred to the UCMC dataset (with the exception of same-admission death). For each outcome, we took the base model and fine-tuned it on the MIMIC training set as described in~\S\ref{ss:sft}. Performance on the test sets is given in Table~\ref{tbl:sft-rocs}. Performance generally improves after finetuning, with particularly pronounced improvements when we transfer these models to UCMC data. Further  ``local finetuning'' on the training split (5\%), with hyperparameter selection on the validation split (5\%) of the UCMC data yielded additional improvement for the ICU admission and IMV event prediction tasks, as seen in Figure~\ref{tbl:internal-sft-rocs}. In analyzing these results it is important to note that the UCMC data are chronologically split where the training period includes patients whose first admission was prior between 1 March 2020 and 5 April 2020. For the validation set, the first admission occurred between 5 April and 5 May 2020, and for the test set, the first admission occurred on or after 5 May 2020.
\begin{table}[ht]
	\centering
	\begin{tabular}{llrrrr}
		\toprule
		dataset                                                                    & subset   & \begin{tabular}[c]{@{}r@{}}same admission\\ death\end{tabular} & \begin{tabular}[c]{@{}r@{}}long length \\ of stay\end{tabular} & ICU admission & IMV event \\ \midrule
		\multirow{3}{*}{\begin{tabular}[c]{@{}l@{}}MIMIC\\ (test)\end{tabular}}    & inliers  & 0.877                                                          & 0.772                                                          & 0.791         & 0.837     \\
		                                                                           & outliers & 0.869                                                          & 0.747                                                          & 0.884         & 0.832     \\
		                                                                           & overall  & 0.902                                                          & 0.783                                                          & 0.795         & 0.853     \\ \midrule
		\multirow{3}{*}{\begin{tabular}[c]{@{}l@{}}UCMC\\ (test)\end{tabular}} & inliers  & 0.856                                                          & 0.652                                                          & 0.556         & 0.615     \\
		                                                                           & outliers & 0.874                                                          & 0.651                                                          & 0.507         & 0.610     \\
		                                                                           & overall  & 0.878                                                          & 0.651                                                          & 0.529         & 0.610     \\ \bottomrule
	\end{tabular}
	\caption{ROC-AUC’s for logistic regression representation-based predictions}
	\label{tbl:rep-based-rocs}
\end{table}

\begin{table}[]
	\centering
	\begin{tabular}{llrrrr}
		\toprule
		dataset                                                                    & subset   & \begin{tabular}[c]{@{}r@{}}same admission\\ death\end{tabular} & \begin{tabular}[c]{@{}r@{}}long length \\ of stay\end{tabular} & ICU admission & IMV event \\ \midrule
		\multirow{3}{*}{\begin{tabular}[c]{@{}l@{}}MIMIC\\ (test)\end{tabular}}    & inliers  & 0.882                                                          & 0.776                                                          & 0.792         & 0.848     \\
		                                                                           & outliers & 0.895                                                          & 0.785                                                          & 0.867         & 0.885     \\
		                                                                           & overall  & 0.910                                                          & 0.789                                                          & 0.795         & 0.867     \\ \midrule
		\multirow{3}{*}{\begin{tabular}[c]{@{}l@{}}UCMC\\ (test)\end{tabular}} & inliers  & 0.905                                                          & 0.740                                                          & 0.606         & 0.653     \\
		                                                                           & outliers & 0.908                                                          & 0.755                                                          & 0.623         & 0.682     \\
		                                                                           & overall  & 0.914                                                          & 0.750                                                          & 0.615         & 0.675     \\ \bottomrule
	\end{tabular}
	\caption{ROC-AUC’s for models fine-tuned on each respective classification task}
	\label{tbl:sft-rocs}
\end{table}

\begin{table}[]
	\centering
	\begin{tabular}{llrrrr}
		\toprule
		dataset                                                                    & subset   & \begin{tabular}[c]{@{}r@{}}same admission\\ death\end{tabular} & \begin{tabular}[c]{@{}r@{}}long length \\ of stay\end{tabular} & ICU admission & IMV event \\ \midrule
		\multirow{3}{*}{\begin{tabular}[c]{@{}l@{}}UCMC\\ (test)\end{tabular}} & inliers  & 0.885                                                          & 0.716                                                          & 0.760         & 0.793     \\
		                                                                           & outliers & 0.883                                                          & 0.733                                                          & 0.767         & 0.811     \\
		                                                                           & overall  & 0.892                                                          & 0.727                                                          & 0.766         & 0.808     \\ \bottomrule
	\end{tabular}
	\caption{ROC-AUC’s for models fine-tuned on MIMIC and then UCMC data}
	\label{tbl:internal-sft-rocs}
\end{table}

\subsection{Representation dynamics and negative outcomes}
\label{s:dynamics}
On the test sets, we used the base model to iteratively extract a sequence of hidden states corresponding to the first 24 hours of each admission sequence. In this way, the final element of each such sequence corresponds to the representation used for the representation-based classifiers and outlier detection. We used the trajectories to calculate path length in representation space and the magnitude of the maximum jump for each hospitalization. We then trained logistic regression models for each outcome of interest using the regressors: path length, max jump, and anomaly score (defined in~\S\ref{ss:out}). In the case of ICU admission and IMV event, we restricted to persons in each dataset who did not experience that outcome within the first 24 hours. For both datasets, each of the regressors had a statistically significant ($p<0.001$) positive-valued effect on both subsequent mortality and long length-of-stay. For MIMIC, max jump and anomaly score had a statistically significant positive-valued effect on subsequent ICU admission and IMV event. For the UCMC data, all three regressors had a highly significant positive-valued effect. For complete results, see appendix section~\ref{ss:logistic_regressions}.

\subsection{Results on application to real-time inpatient mortality predictions in the first 24 hours}
\begin{figure}[!pht]
	\centering
	\subfigure[SFT predictions on MIMIC]{
		\centering
		\includegraphics[width=0.43\textwidth]{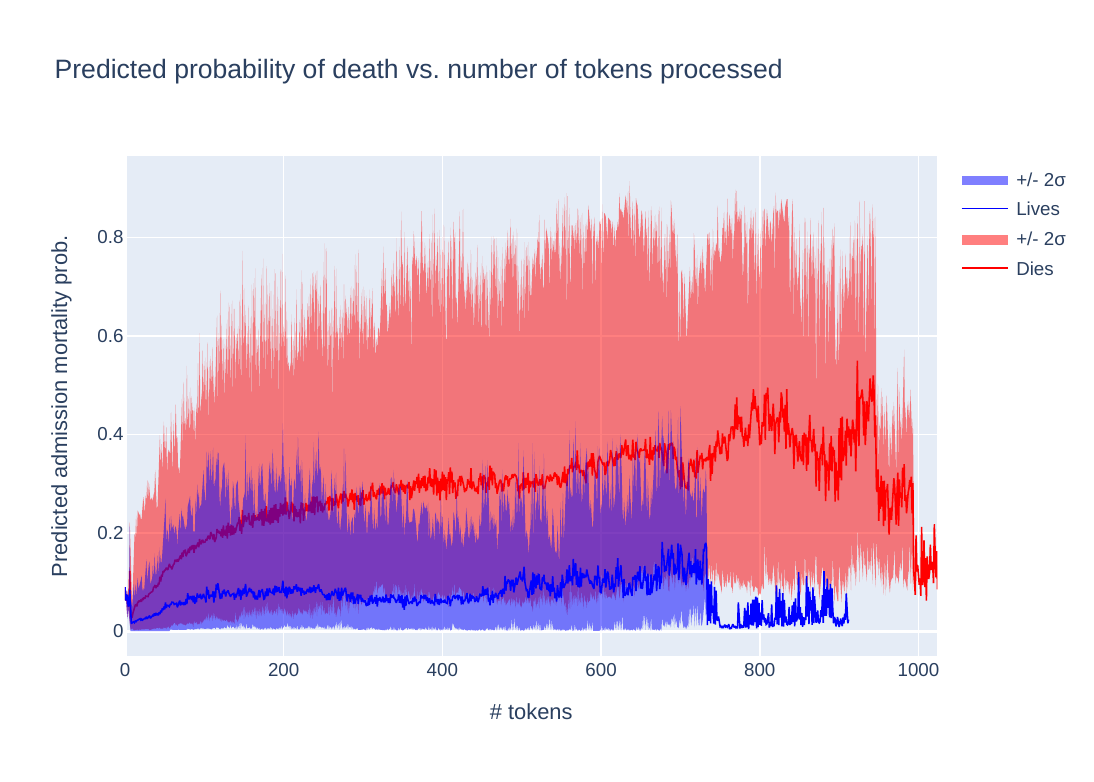}
		\label{fig:mimic_all}
	}
	\hspace{10pt}
	\subfigure[SFT predictions on UCMC]{
		\centering
		\includegraphics[width=0.43\textwidth]{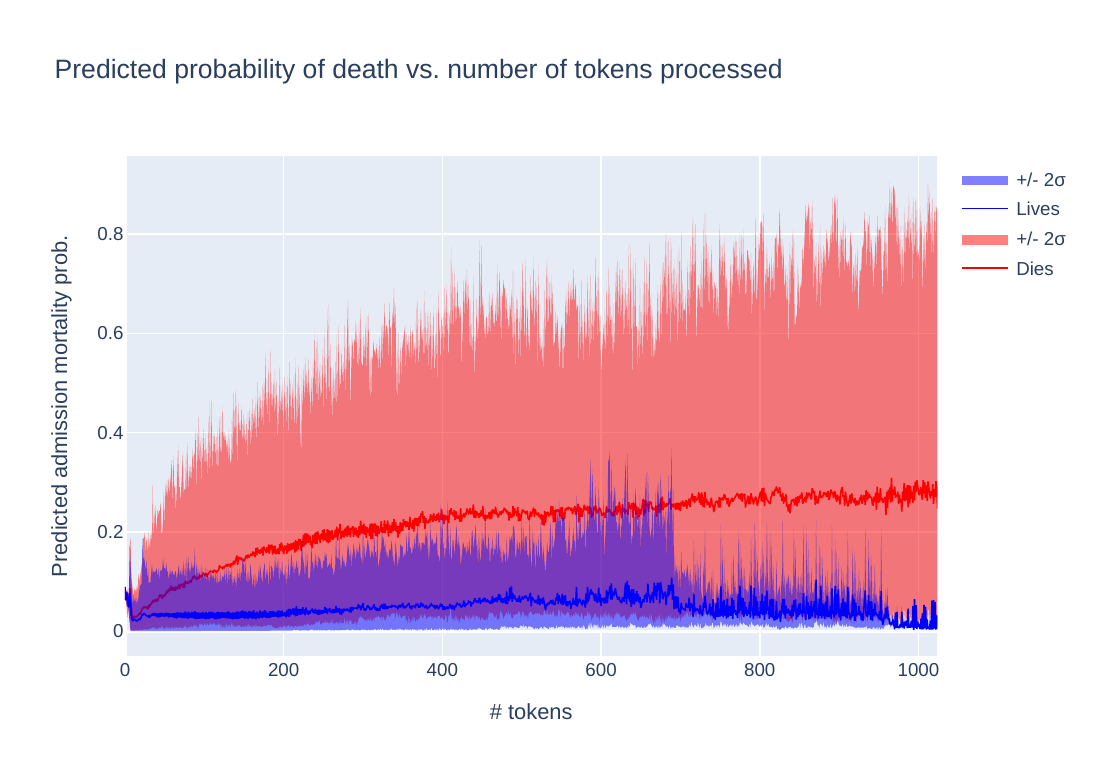}
		\label{fig:internal_all}
	}
	\subfigure[SFT w/ URT on MIMIC]{
		\centering
		\includegraphics[width=0.43\textwidth]{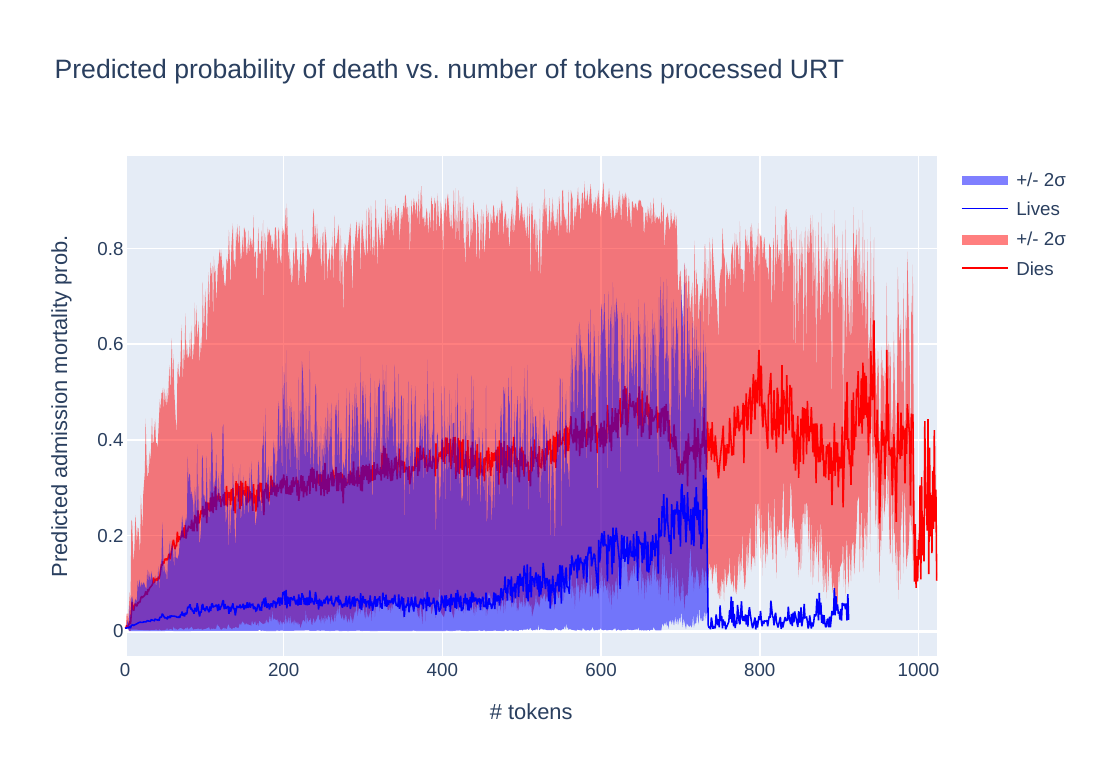}
		\label{fig:urt_mimic_all}
	}
	\hspace{10pt}
	\subfigure[SFT w/ URT on UCMC]{
		\centering
		\includegraphics[width=0.43\textwidth]{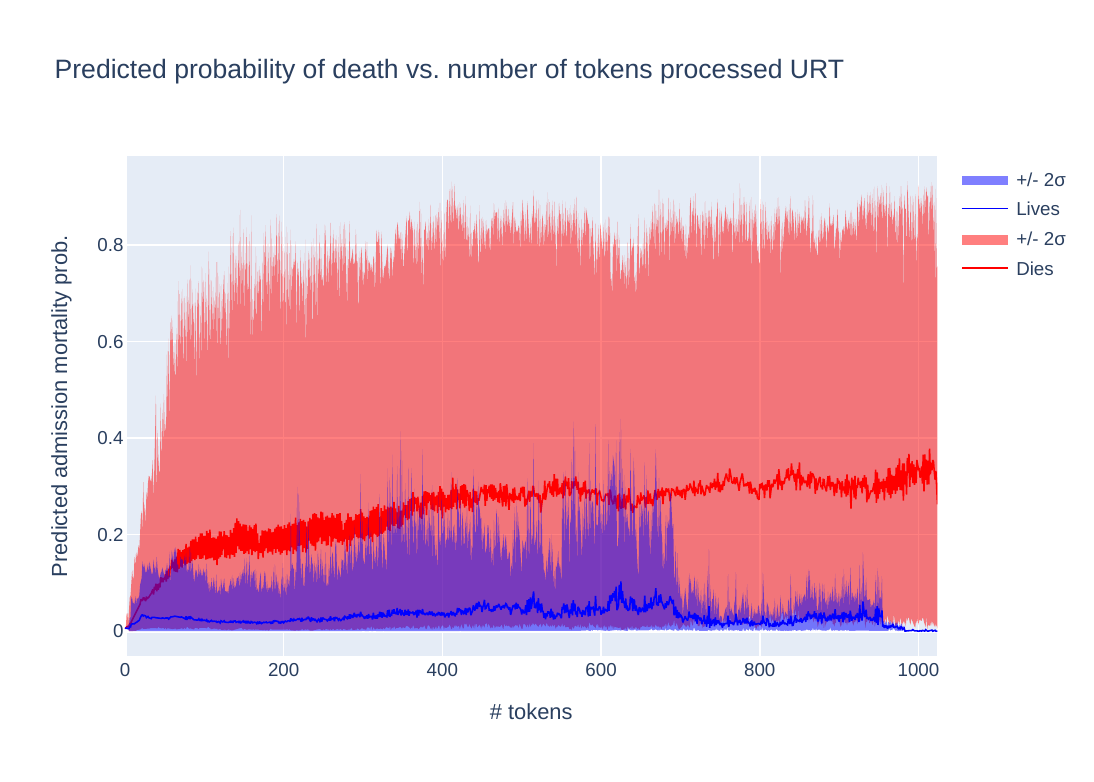}
		\label{fig:urt_internal_all}
	}
	\subfigure[LR w/ URT on MIMIC]{
		\centering
		\includegraphics[width=0.43\textwidth]{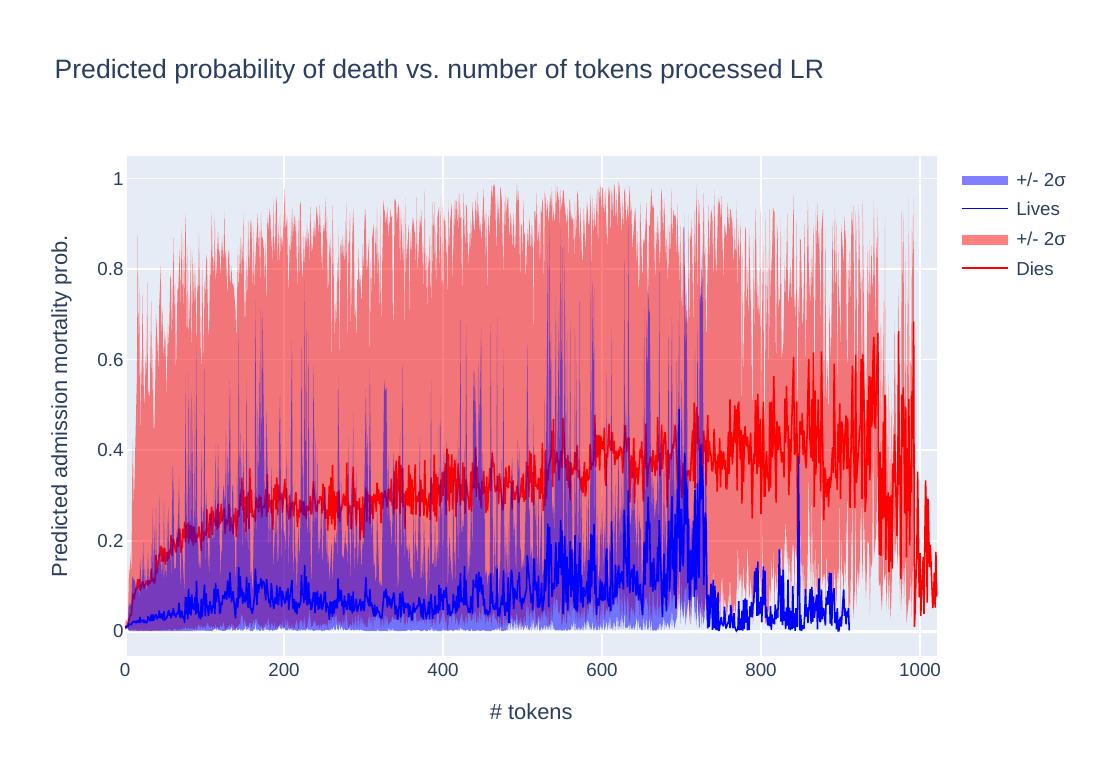}
		\label{fig:lr_mimic_all}
	}
	\hspace{10pt}
	\subfigure[LR w/ URT on UCMC]{
		\centering
		\includegraphics[width=0.43\textwidth]{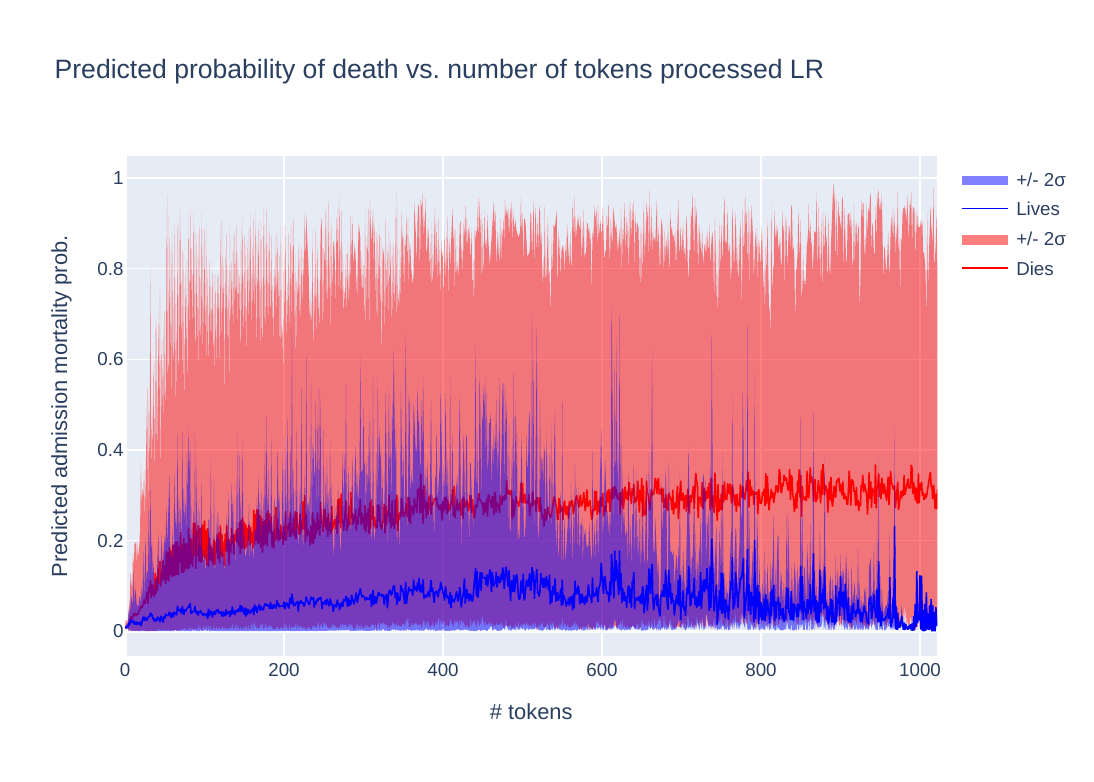}
		\label{fig:lr_internal_all}
	}
	\caption{For 100 timelines corresponding to inpatient mortality and for 100 timelines that do not, we plot the mean and a 95\% quantile range for inpatient mortality predictions for the first $i$ tokens. On the left, we have results for MIMIC and on the right are results for UCMC. Subfigures ({\it a}) and ({\it b}) correspond to predictions from supervised fine-tuning (SFT) on the MIMIC training sequences on ({\it a}) the MIMIC test set and ({\it b}) the UCMC test set. Subfigures ({\it c}) and ({\it d}) correspond to predictions from supervised fine-tuning (SFT) on MIMIC training sequences with uniform random truncation (URT). Subfigures ({\it e}) and ({\it f}) correspond to predictions from logistic regression classifiers trained on representations from the original model (no fine-tuning) extracted from sequences with uniform random truncation.}
	\label{fig:real_time_preds}
\end{figure}

For both the MIMIC and UCMC datasets, we sampled 100 patients uniformly at random from those with outcomes corresponding to inpatient mortality and those who survived their hospitalization. We then applied the 3 classifiers for partial sequence prediction as developed in~\S\ref{ss:partial_seq_clsf} over the entire extent of each timeline up to the 24-hour cutoff point. See Figure~\ref{fig:real_time_preds} for aggregated real-time predictions. Introducing truncated sequences into the training data may allow us to identify patients at risk of death earlier in their timelines.

\section{Discussion}

Our study provides insights into the behavior and utility of foundation models for electronic health records, particularly in the context of transferability across healthcare systems. While representations trained on pre-pandemic patient data from Beth Israel Deaconess Medical Center (MIMIC-IV) performed well within-distribution, we observed substantial degradation in performance when applied to a test dataset from UCMC collected during the COVID-19 pandemic, particularly for outcomes such as ICU admission and IMV events. This geographic and temporal mismatch between the training and test datasets setting provides a realistic and challenging testbed for model transferability. The substantial domain shift reflects differences in case mix and resource availability, but also evolving clinical protocols, triage policies, and documentation patterns. These differences likely explain the poor cross-site generalization for outcomes like ICU admission and IMV events when not performing local finetuning and highlight the challenges of model transfer in healthcare settings.

Inpatient mortality prediction demonstrated a far greater robustness between health systems. Mortality is a highly distinct and well-documented event. As such, it may be easier for the FM to learn robust cross-institutional patterns associated with imminent death, even in the presence of systemic distribution shifts. In contrast, ICU transfer and IMV initiation are more susceptible to local hospital policies, bed availability, and pandemic-era triage procedures~\citep{Pis20}, factors that may not be directly observable in the data but likely affect model performance.

Fine-tuning the foundation model substantially improved performance, particularly for the UCMC COVID-19 pandemic era test dataset. This suggests that even when a substantial portion of the target dataset appears anomalous relative to the source distribution (nearly half of our COVID-19 pandemic era test samples were classified as outliers), supervised fine-tuning can effectively adapt the model. The additional benefit observed from ``local fine-tuning'' on a small COVID-19 pandemic era training set (5\%) for ICU admission and IMV event prediction indicates that even limited target-domain data can significantly enhance performance for certain tasks but some performance gaps remained. This was true even though the target-domain data used for finetuning (first 5\%) captured the beginning of the COVID-19 pandemic (March 2020 to April 2020) where there was great uncertainty and substantial disruptions to hospital operations. The UCMC validation and test sets covered the duration of the pandemic, where there were varied conditions with pandemic surges and normal operations. This variation in hospital strain, practice standards, and outcomes represents a uniquely difficult challenge for transferability.

Our analysis of representation dynamics revealed consistent patterns across both datasets: trajectory path length, maximum jump magnitude, and anomaly scores all showed positive associations with adverse outcomes. This consistency suggests that despite differences in data distributions, the way patient trajectories evolve in representation space captures clinically meaningful information that generalizes across healthcare institutions, diagnoses, and time. The fact that these representation-derived features strongly predict future clinical deterioration even before it becomes apparent in clinical practice highlights the potential value of these models for early risk stratification and calls for deeper study of the dynamics of patient trajectories through the learned clinical latent space.

The evaluation of partial sequence prediction demonstrates that fine-tuned models can identify high-risk patients earlier in their hospital course, with stronger discrimination between mortality outcomes occurring as more information becomes available. However, the high variability in predictions observed, particularly in later time points for the MIMIC dataset, suggests limitations in prediction stability that warrant further investigation.

Our results collectively emphasize that while EHR FMs offer significant promise, their successful deployment requires careful consideration of transferability and adaptation. Fine-tuning, potentially augmented by local data, appears essential for bridging the gap between different healthcare systems. Moreover, analyzing the dynamics of learned representations can offer valuable insights into model behavior and patient risk.

\subsection{Limitations}
Several limitations of this study should be noted. First, some of the degradation in the performance of the FM could be secondary to imperfect data standardization. While the CLIF consortium aims to reduce data disparities between member hospitals by harmonizing ICU-relevant data elements across institutions, full semantic alignment is difficult to guarantee. Differences in the frequency and granularity of measurements, variation in device labeling, and heterogeneous use of specific data tables (e.g., respiratory support, assessments) may introduce noise into token sequences and model inputs. Our decision to encode labs and medications at a broad categorical level significantly reduced the overall vocabulary, allowing more efficient training with smaller data. However, it can merge clinically distinct items into the same token. For instance, potassium measured from whole blood versus serum is treated identically, obscuring an important factor for distinguishing pseudohyperkalemia from life-threatening hyperkalemia. Although each such scenario may be rare, the cumulative effect of losing these finer distinctions can weaken a model's capacity to capture certain clinical signals.

Second, certain common clinically important labs have not yet been standardized into CLIF categories; for example, thyroid-related tests that might detect myxedema coma or thyroid storm are not yet included. While focusing initially on a smaller vocabulary helps standardize data across sites, it can also limit the model's granularity and miss key markers. Furthermore, this current work addresses only MIMIC-IV and UCMC, but CLIF provides an opportunity to evaluate foundational model transferability and patient representation dynamics across 8 additional health systems.

Finally, our anomaly detection approach, while interpretable and computationally efficient, is relatively simple. Future work could explore more expressive unsupervised methods, including generative models or density estimation approaches that better capture the dynamic nature of patient state representations. Tied to this, we currently restrict the input data to a 24-hour observation window. From a prediction task perspective, this represents a realistic utility, but we expect that longer longitudinal timelines may present even greater opportunities for trajectory analyses. We plan to expand to longer time horizons, and study how model representations evolve across multiple hospitalizations or chronic disease courses.

\acks{This work was funded in part by the National Institutes of Health, specifically the National Institute of Neurological Disorders and Stroke grant number R00NS114850 to BKB. This project would not have been possible without the support of the Center for Research Informatics at the University of Chicago and particularly the High-Performance Computing team led by Mike Jarsulic. The authors are grateful for the resources and support this team provided throughout the duration of the project. The Center for Research Informatics is funded by the Biological Sciences Division at the University of Chicago with additional funding provided by the Institute for Translational Medicine, CTSA grant number UL1 TR000430 from the National Institutes of Health.}


\newpage
\appendix
\section{Code supplement}
\label{ss:code}

\begin{figure}[ht]
	\centering
	\includegraphics[width=\textwidth]{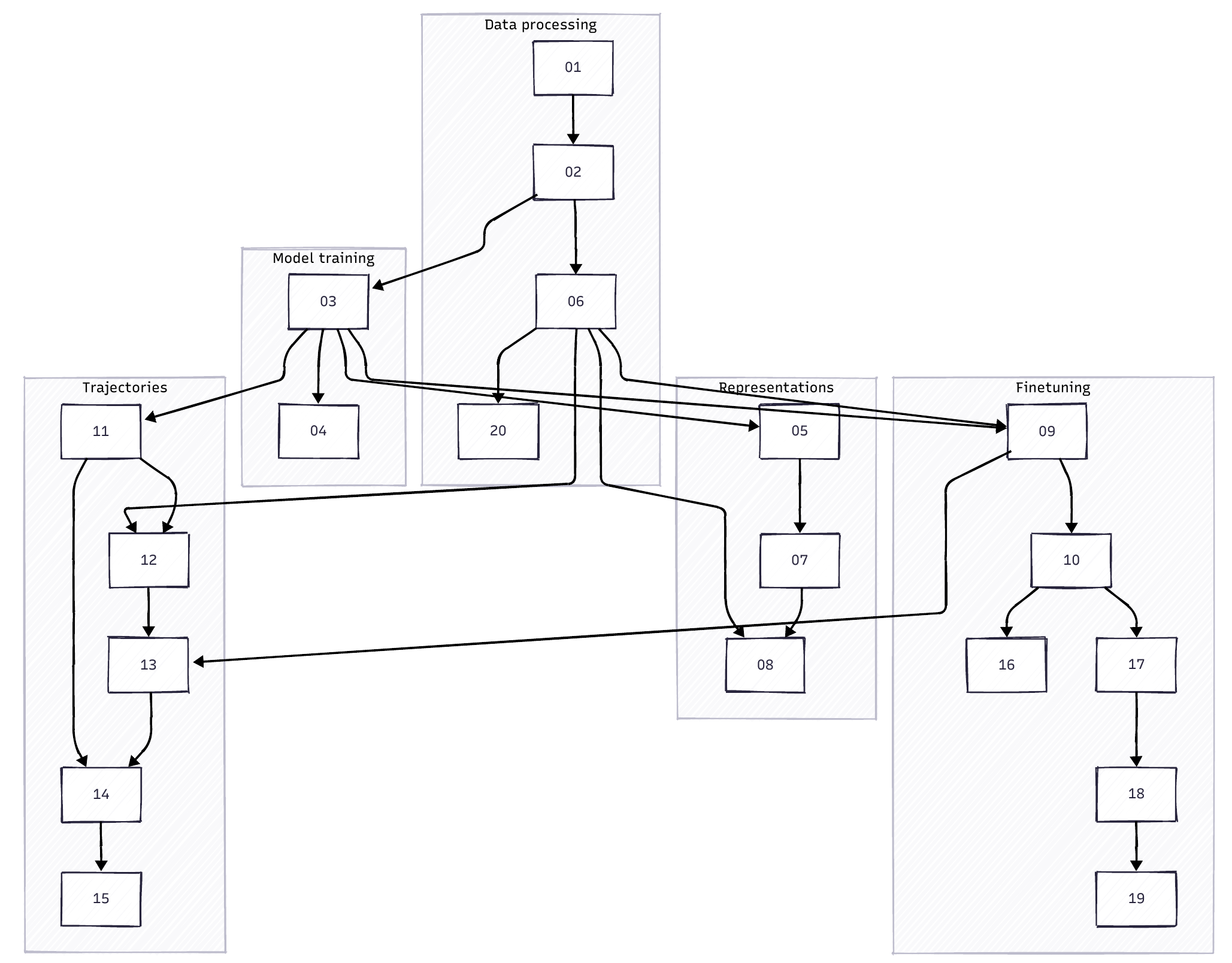}
	\caption{Our code is organized logically as shown above. Running the provided slurm scripts in this order (with access to compute nodes containing 8 Nvidia A100 GPUs with two 16-core 3.0-GHz AMD Milan processors) with the provided requirements file on MIMIC data and the UCMC dataset converted to the CLIF format produces the results contained in this manuscript.}
	\label{fig:code}
\end{figure}

The code necessary to reproduce the results provided in this manuscript is publicly available on Github:

\url{https://github.com/bbj-lab/FMs-EHRs-Rep-Dynamics-and-Transfer}\\
Figure~\ref{fig:code} describes the logical flow of the provided scripts.

\section{Results from logistic regressions for outcomes against trajectory properties and anomaly scores}
\label{ss:logistic_regressions}

In this appendix section, we provide results from the logistic regression analysis described in~\S\ref{s:dynamics}. Tables~\ref{t:mimic_lr_mort}--\ref{t:mimic_lr_imve} provide results on the MIMIC test set for inpatient mortality, long length of stay, ICU admission (after 24 hours given no admission in the first 24 hours), and IMV event (again, after 24 hours given no event in the first 24 hours). Tables~\ref{t:internal_lr_mort}--\ref{t:internal_lr_imve} provide analogous results for the UCMC test set. The third and fourth outcomes entail restrictions; this is why the number of observations decreases for these models.

\begin{table}[!ht]
	\begin{center}
		\begin{tabular}{lclc}
			\toprule
			\textbf{Dep. Variable:}        & same\_admission\_death & \textbf{  No. Observations:  } & 90240   \\
			\textbf{Model:}                & Logit                  & \textbf{  Df Residuals:      } & 90236   \\
			\textbf{Method:}               & MLE                    & \textbf{  Df Model:          } & 3       \\
			\textbf{  Pseudo R-squ.:     } & 0.1443                 & \textbf{  LL-Null:           } & -9664.4 \\
			\textbf{  Log-Likelihood:    } & -8269.8                & \textbf{  LLR p-value:       } & 0.000   \\
			\midrule
		\end{tabular}
		\begin{tabular}{lcccccc}
			                           & \textbf{coef} & \textbf{std err} & \textbf{z} & \textbf{P$> |$z$|$} & \textbf{[0.025} & \textbf{0.975]} \\
			\midrule
			\textbf{Intercept}         & -30.2686      & 1.816            & -16.664    & 0.000               & -33.829         & -26.709         \\
			\textbf{Trajectory Length} & 4.839e-05     & 2.58e-06         & 18.748     & 0.000               & 4.33e-05        & 5.34e-05        \\
			\textbf{Maximum Jump}      & 0.3876        & 0.031            & 12.613     & 0.000               & 0.327           & 0.448           \\
			\textbf{Anomaly Score}     & 6.9725        & 0.666            & 10.466     & 0.000               & 5.667           & 8.278           \\
			\bottomrule
		\end{tabular}
		\caption{Logistic regression results for inpatient mortality on the MIMIC test set}
		\label{t:mimic_lr_mort}
	\end{center}
\end{table}

\begin{table}[!ht]
	\begin{center}
		\begin{tabular}{lclc}
			\toprule
			\textbf{Dep. Variable:}        & long\_length\_of\_stay & \textbf{  No. Observations:  } & 90240   \\
			\textbf{Model:}                & Logit                  & \textbf{  Df Residuals:      } & 90236   \\
			\textbf{Method:}               & MLE                    & \textbf{  Df Model:          } & 3       \\
			\textbf{  Pseudo R-squ.:     } & 0.06410                & \textbf{  LL-Null:           } & -47555. \\
			\textbf{  Log-Likelihood:    } & -44506.                & \textbf{  LLR p-value:       } & 0.000   \\
			\midrule
		\end{tabular}
		\begin{tabular}{lcccccc}
			                           & \textbf{coef} & \textbf{std err} & \textbf{z} & \textbf{P$> |$z$|$} & \textbf{[0.025} & \textbf{0.975]} \\
			\midrule
			\textbf{Intercept}         & -32.9854      & 0.711            & -46.420    & 0.000               & -34.378         & -31.593         \\
			\textbf{Trajectory Length} & 1.907e-05     & 1.31e-06         & 14.595     & 0.000               & 1.65e-05        & 2.16e-05        \\
			\textbf{Maximum Jump}      & 0.5030        & 0.012            & 42.658     & 0.000               & 0.480           & 0.526           \\
			\textbf{Anomaly Score}     & 4.5664        & 0.266            & 17.150     & 0.000               & 4.044           & 5.088           \\
			\bottomrule
		\end{tabular}
		\caption{Logistic regression results for long length of stay on the MIMIC test set}
	\end{center}
\end{table}

\begin{table}[!h]
	\begin{center}
		\begin{tabular}{lclc}
			\toprule
			\textbf{Dep. Variable:}        & icu\_admission & \textbf{  No. Observations:  } & 77135      \\
			\textbf{Model:}                & Logit          & \textbf{  Df Residuals:      } & 77131      \\
			\textbf{Method:}               & MLE            & \textbf{  Df Model:          } & 3          \\
			\textbf{  Pseudo R-squ.:     } & 0.04798        & \textbf{  LL-Null:           } & -14195.    \\
			\textbf{  Log-Likelihood:    } & -13514.        & \textbf{  LLR p-value:       } & 4.845e-295 \\
			\midrule
		\end{tabular}
		\begin{tabular}{lcccccc}
			                           & \textbf{coef} & \textbf{std err} & \textbf{z} & \textbf{P$> |$z$|$} & \textbf{[0.025} & \textbf{0.975]} \\
			\midrule
			\textbf{Intercept}         & -55.8534      & 1.778            & -31.408    & 0.000               & -59.339         & -52.368         \\
			\textbf{Trajectory Length} & -3.515e-05    & 7.85e-06         & -4.479     & 0.000               & -5.05e-05       & -1.98e-05       \\
			\textbf{Maximum Jump}      & 0.8211        & 0.029            & 28.428     & 0.000               & 0.764           & 0.878           \\
			\textbf{Anomaly Score}     & 10.8883       & 0.615            & 17.713     & 0.000               & 9.683           & 12.093          \\
			\bottomrule
		\end{tabular}
		\caption{Logistic regression results for ICU admission on the MIMIC test set}
	\end{center}
\end{table}

\begin{table}[!h]
	\begin{center}
		\begin{tabular}{lclc}
			\toprule
			\textbf{Dep. Variable:}        & imv\_event & \textbf{  No. Observations:  } & 86422   \\
			\textbf{Model:}                & Logit      & \textbf{  Df Residuals:      } & 86418   \\
			\textbf{Method:}               & MLE        & \textbf{  Df Model:          } & 3       \\
			\textbf{  Pseudo R-squ.:     } & 0.08366    & \textbf{  LL-Null:           } & -10818. \\
			\textbf{  Log-Likelihood:    } & -9912.7    & \textbf{  LLR p-value:       } & 0.000   \\
			\midrule
		\end{tabular}
		\begin{tabular}{lcccccc}
			                           & \textbf{coef} & \textbf{std err} & \textbf{z} & \textbf{P$> |$z$|$} & \textbf{[0.025} & \textbf{0.975]} \\
			\midrule
			\textbf{Intercept}         & -41.1799      & 1.670            & -24.666    & 0.000               & -44.452         & -37.908         \\
			\textbf{Trajectory Length} & -1.432e-06    & 3.09e-06         & -0.463     & 0.643               & -7.5e-06        & 4.63e-06        \\
			\textbf{Maximum Jump}      & 0.5369        & 0.028            & 19.176     & 0.000               & 0.482           & 0.592           \\
			\textbf{Anomaly Score}     & 13.7308       & 0.579            & 23.696     & 0.000               & 12.595          & 14.867          \\
			\bottomrule
		\end{tabular}
		\caption{Logistic regression results for IMV event on the MIMIC test set}
		\label{t:mimic_lr_imve}
	\end{center}
\end{table}

\begin{table}[!h]
	\begin{center}
		\begin{tabular}{lclc}
			\toprule
			\textbf{Dep. Variable:}        & same\_admission\_death & \textbf{  No. Observations:  } & 53991   \\
			\textbf{Model:}                & Logit                  & \textbf{  Df Residuals:      } & 53987   \\
			\textbf{Method:}               & MLE                    & \textbf{  Df Model:          } & 3       \\
			\textbf{  Pseudo R-squ.:     } & 0.2208                 & \textbf{  LL-Null:           } & -6679.0 \\
			\textbf{  Log-Likelihood:    } & -5204.0                & \textbf{  LLR p-value:       } & 0.000   \\
			\midrule
		\end{tabular}
		\begin{tabular}{lcccccc}
			                           & \textbf{coef} & \textbf{std err} & \textbf{z} & \textbf{P$> |$z$|$} & \textbf{[0.025} & \textbf{0.975]} \\
			\midrule
			\textbf{Intercept}         & -85.8206      & 3.469            & -24.739    & 0.000               & -92.620         & -79.021         \\
			\textbf{Trajectory Length} & 5.628e-05     & 2.42e-06         & 23.273     & 0.000               & 5.15e-05        & 6.1e-05         \\
			\textbf{Maximum Jump}      & 1.3095        & 0.058            & 22.451     & 0.000               & 1.195           & 1.424           \\
			\textbf{Anomaly Score}     & 6.6050        & 0.859            & 7.688      & 0.000               & 4.921           & 8.289           \\
			\bottomrule
		\end{tabular}
		\caption{Logistic regression results for inpatient mortality on the UCMC test set}
		\label{t:internal_lr_mort}
	\end{center}
\end{table}

\begin{table}[!h]
	\begin{center}
		\begin{tabular}{lclc}
			\toprule
			\textbf{Dep. Variable:}        & long\_length\_of\_stay & \textbf{  No. Observations:  } & 53991   \\
			\textbf{Model:}                & Logit                  & \textbf{  Df Residuals:      } & 53987   \\
			\textbf{Method:}               & MLE                    & \textbf{  Df Model:          } & 3       \\
			\textbf{  Pseudo R-squ.:     } & 0.08189                & \textbf{  LL-Null:           } & -30580. \\
			\textbf{  Log-Likelihood:    } & -28076.                & \textbf{  LLR p-value:       } & 0.000   \\
			\midrule
		\end{tabular}
		\begin{tabular}{lcccccc}
			                           & \textbf{coef} & \textbf{std err} & \textbf{z} & \textbf{P$> |$z$|$} & \textbf{[0.025} & \textbf{0.975]} \\
			\midrule
			\textbf{Intercept}         & -59.4728      & 1.225            & -48.546    & 0.000               & -61.874         & -57.072         \\
			\textbf{Trajectory Length} & 2.263e-05     & 9.51e-07         & 23.786     & 0.000               & 2.08e-05        & 2.45e-05        \\
			\textbf{Maximum Jump}      & 0.9597        & 0.020            & 46.848     & 0.000               & 0.920           & 1.000           \\
			\textbf{Anomaly Score}     & 2.8881        & 0.378            & 7.641      & 0.000               & 2.147           & 3.629           \\
			\midrule
		\end{tabular}
		\caption{Logistic regression results for long length of stay on the UCMC test set}
	\end{center}
\end{table}

\begin{table}[!h]
	\begin{center}
		\begin{tabular}{lclc}
			\toprule
			\textbf{Dep. Variable:}        & icu\_admission & \textbf{  No. Observations:  } & 46219      \\
			\textbf{Model:}                & Logit          & \textbf{  Df Residuals:      } & 46215      \\
			\textbf{Method:}               & MLE            & \textbf{  Df Model:          } & 3          \\
			\textbf{  Pseudo R-squ.:     } & 0.04518        & \textbf{  LL-Null:           } & -8932.3    \\
			\textbf{  Log-Likelihood:    } & -8528.8        & \textbf{  LLR p-value:       } & 1.222e-174 \\
			\midrule
		\end{tabular}
		\begin{tabular}{lcccccc}
			                           & \textbf{coef} & \textbf{std err} & \textbf{z} & \textbf{P$> |$z$|$} & \textbf{[0.025} & \textbf{0.975]} \\
			\midrule
			\textbf{Intercept}         & -61.5467      & 2.567            & -23.972    & 0.000               & -66.579         & -56.515         \\
			\textbf{Trajectory Length} & 2.428e-05     & 2.29e-06         & 10.588     & 0.000               & 1.98e-05        & 2.88e-05        \\
			\textbf{Maximum Jump}      & 0.9318        & 0.042            & 21.993     & 0.000               & 0.849           & 1.015           \\
			\textbf{Anomaly Score}     & 6.6612        & 0.862            & 7.725      & 0.000               & 4.971           & 8.351           \\
			\midrule
		\end{tabular}
		\caption{Logistic regression results for ICU admission on the UCMC test set}
	\end{center}
\end{table}

\begin{table}[!h]
	\begin{center}
		\begin{tabular}{lclc}
			\toprule
			\textbf{Dep. Variable:}        & imv\_event & \textbf{  No. Observations:  } & 51350      \\
			\textbf{Model:}                & Logit      & \textbf{  Df Residuals:      } & 51346      \\
			\textbf{Method:}               & MLE        & \textbf{  Df Model:          } & 3          \\
			\textbf{  Pseudo R-squ.:     } & 0.08560    & \textbf{  LL-Null:           } & -6179.0    \\
			\textbf{  Log-Likelihood:    } & -5650.1    & \textbf{  LLR p-value:       } & 5.001e-229 \\
			\midrule
		\end{tabular}
		\begin{tabular}{lcccccc}
			                           & \textbf{coef} & \textbf{std err} & \textbf{z} & \textbf{P$> |$z$|$} & \textbf{[0.025} & \textbf{0.975]} \\
			\midrule
			\textbf{Intercept}         & -73.3559      & 3.303            & -22.210    & 0.000               & -79.829         & -66.882         \\
			\textbf{Trajectory Length} & 3.495e-05     & 2.45e-06         & 14.291     & 0.000               & 3.02e-05        & 3.97e-05        \\
			\textbf{Maximum Jump}      & 1.1174        & 0.055            & 20.334     & 0.000               & 1.010           & 1.225           \\
			\textbf{Anomaly Score}     & 6.2521        & 0.990            & 6.314      & 0.000               & 4.311           & 8.193           \\
			\bottomrule
		\end{tabular}
		\caption{Logistic regression results for IMV event on the UCMC test set}
		\label{t:internal_lr_imve}
	\end{center}
\end{table}

\end{document}